%% file: main.tex
\documentclass[10pt,twocolumn,letterpaper]{article}

\usepackage{iccv}              % To produce the CAMERA-READY version
\usepackage{graphicx}
\usepackage{multirow}
\usepackage{pifont}
\usepackage{graphicx}
\usepackage{colortbl}
\usepackage{latexsym}

\definecolor{iccvblue}{rgb}{0.21,0.49,0.74}
\usepackage[pagebackref,breaklinks,colorlinks,allcolors=iccvblue]{hyperref}
\usepackage[accsupp]{axessibility}  % Improves PDF readability for those with disabilities.

\def\paperID{2497} % *** Enter the Paper ID here
\def\confName{ICCV}
\def\confYear{2025}

\begin{document}
\title{SpatialSplat: Efficient Semantic 3D from Sparse Unposed Images}

%%%%%%%%% AUTHORS - PLEASE UPDATE
\author{
Yu Sheng\textsuperscript{1}, 
Jiajun Deng\textsuperscript{2}$^\dag$, 
Xinran Zhang\textsuperscript{1}, 
Yu Zhang\textsuperscript{1}, 
Bei Hua\textsuperscript{1}, 
Yanyong Zhang\textsuperscript{1}, 
Jianmin Ji\textsuperscript{1}$^\dag$\\
$^{1}$University of Science and Technology of China, $^{2}$The University of Adelaide\\
{\tt\small \{shengyu724,zxrr\}@mail.ustc.edu.cn, jiajun.deng@adelaide.edu.au}\\
{\tt\small \{zhangyu,bhua,yanyongz,jianmin\}@ustc.edu.cn} \quad
{\small $\dag$: Corresponding Author}
}
% For a paper whose authors are all at the same institution,
% omit the following lines up until the closing ``}''.
% Additional authors and addresses can be added with ``\and'',
% just like the second author.
% To save space, use either the email address or home page, not both
% \and
% Second Author\\
% Institution2\\
% First line of institution2 address\\
% {\tt\small secondauthor@i2.org}

\twocolumn[{%
\renewcommand\twocolumn[1][]{#1}%
\maketitle
\begin{center}
    \centering
    \includegraphics[width=.998\textwidth]{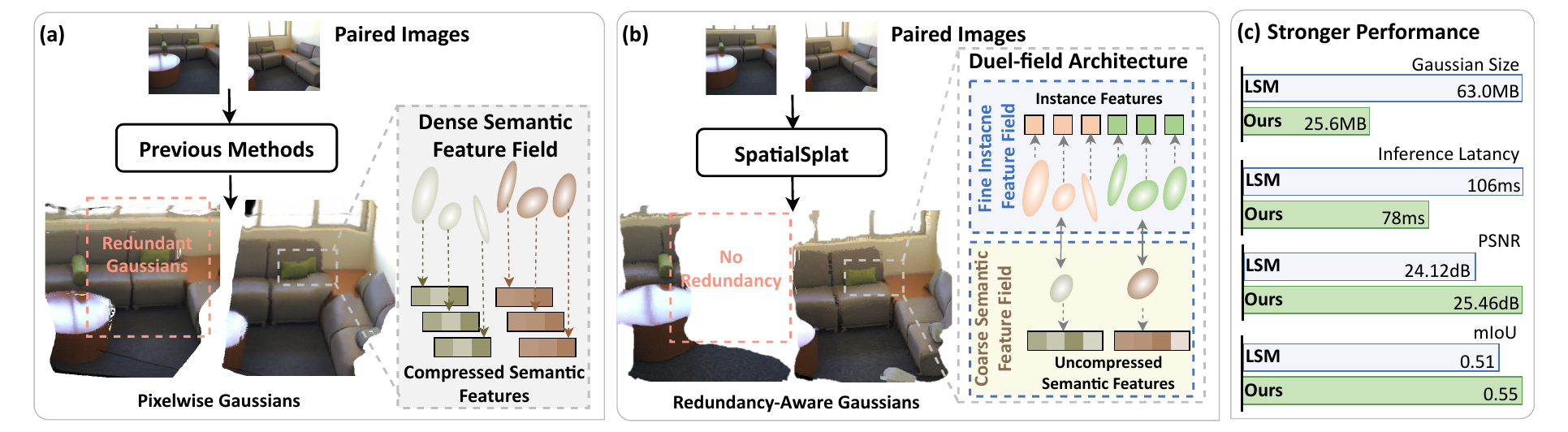}
    \captionof{figure}{\textbf{Comparison between previous methods and our SpatialSplat.} 
    (a): Previous methods predict pixel-wise Gaussians, associating each primitive with compressed semantic feature.
    (b): Our SpatialSplat avoids generating redundant Gaussian primitives for overlapping pixels, and it represents semantics with a dual-field architecture to better preserve the information.
    (c): SpatialSplat outperforms the state-of-the-art method LSM~\cite{lsm} for both novel-view rendering quality and semantic segmentation while being more efficient.}
    \label{fig:first-fig}
\end{center}%
}]

\newcommand{\snote}[1] {{$\langle${\textcolor{red}{sheng: \textbf{#1}}}$\rangle$}}
\newcommand{\xnote}[1] {{$\langle${\textcolor{blue}{Xinran: \textbf{#1}}}$\rangle$}}

\definecolor{myblue}{RGB}{228, 246, 253}

\input{sec/0_abstract}    
\input{sec/1_intro}
\input{sec/2_related_work}
\input{sec/3_method}
\input{sec/4_exp}
\input{sec/5_conclusion}

\input{sec/Acknowledgments}
{
    \small
    \bibliographystyle{ieeenat_fullname}
    \bibliography{main}
}

% WARNING: do not forget to delete the supplementary pages from your submission 
% \appendix
% \input{sec/X_suppl}

\end{document}

%% file: sec/0_abstract.tex
\begin{abstract}

A major breakthrough in 3D reconstruction is the feedforward paradigm to generate pixel-wise 3D points or Gaussian primitives from sparse, unposed images. To further incorporate semantics while avoiding the significant memory and storage costs of high-dimensional semantic features, existing methods extend this paradigm by associating each primitive with a compressed semantic feature vector.
However, these methods have two major limitations: (a) the naively compressed feature compromises expressiveness, affecting the model's ability to capture fine-grained semantics, and (b) the pixel-wise primitive prediction introduces redundancy in overlapping areas, causing unnecessary memory overhead.
To this end, we introduce \textbf{SpatialSplat}, a feedforward framework that produces redundancy-aware Gaussians and capitalizes on a dual-field semantic representation. 
Particularly, with the insight that primitives within the same instance exhibit high semantic consistency, we decompose the semantic representation into a coarse feature field that encodes uncompressed semantics with minimal primitives, and a fine-grained yet low-dimensional feature field that captures detailed inter-instance relationships.
Moreover, we propose a selective Gaussian mechanism, which retains only essential Gaussians in the scene, effectively eliminating redundant primitives.
Our proposed Spatialsplat learns accurate semantic information and detailed instances prior with more compact 3D Gaussians, making semantic 3D reconstruction more applicable.
We conduct extensive experiments to evaluate our method, demonstrating a remarkable 60\% reduction in scene representation parameters while achieving superior performance over state-of-the-art methods. The code is available at \href{https://github.com/shengyuuu/SpatialSplat.git}{\textbf{SpatialSplat}}.

\end{abstract}

%% file: sec/1_intro.tex
\section{Introduction}
\label{sec:intro}
% reconstruction important -> semantic
% reconstruction + semantic is not t
Reconstructing and understanding 3D scenes from 2D images~\cite{VR,AR,GEFF,F3RM,survey} is a fundamental topic in computer vision, aiming to obtain semantic-aware 3D structure from low-cost devices, \emph{i.e.}, RGB cameras. This technique is significant for various applications, such as robotics, autonomous driving, and VR/AR.
Recently, with the emergence of the new 3D representation, \emph{i.e.}, Neural Radiance Fields (NeRF)~\cite{NERF} and 3D Gaussian Splatting (3DGS)~\cite{3dgs}, a popular paradigm of performing semantic-aware 3D reconstruction is to distill the feature field of NeRF or 3DGS with powerful 2D vision-language foundation models~\cite{lerf,Feature3dgs,langsplat,Feature-platting}.
% Recent advances~\cite{NERF,3dgs,Feature-platting,lsm} integrate vision-language models with powerful representations like neural radiance fields (NeRF)~\cite{NERF} and 3D Gaussian Splatting (3DGS)~\cite{3dgs} through feature field distillation, enabling simultaneous scene reconstruction and understanding.
However, these methods typically rely on per-scene optimization and complex multi-step preprocessing, 
limiting their ability to generalize across multiple scenes within a single model.

Recent breakthroughs~\cite{mvsplat,pixel-splat,transsplat} in 3D reconstruction have incorporated feed-forward networks to improve the generalization of 3DGS models and accelerate reconstruction.
These methods follow a paradigm that generates pixel-wise Gaussian primitives from sparse posed images.
Building on this, further efforts~\cite{ye2024noposplat,smart2024splatt3r} have extended the paradigm to reconstruct scenes from unposed images.
Moreover, methods~\cite{lsm,GSemSplat} have advanced this approach by integrating semantic understanding, enabling semantic 3D reconstruction from sparse unposed images.

Despite significant progress, these methods have two major limitations.
First, as shown in Fig.\ref{fig:first-fig} (a), pixel-wise Gaussian prediction introduces substantial redundancy in overlapping regions while providing minimal accuracy gains.
Method~\cite{free-spalt} attempts to mitigate this issue by projecting primitives onto a plane and identifying overlaps, it depends on precise camera extrinsics, which are often difficult to obtain in real-world scenarios—especially when only a few sparse, textureless images are available.
Second, methods~\cite{Feature3dgs,langsplat,Feature-platting,GSemSplat,lsm} enforce per-primitive semantic learning to form a dense semantic feature field, as shown in Fig.\ref{fig:first-fig} (a).
This primitive-level approach incurs significant memory and storage costs due to high-dimensional (512 or more) language features.
To mitigate this, the methods often compress features into lower-dimensional representations (e.g., 64-128), but this compression leads to irreversible information loss~\cite{Feature3dgs}, leading to suboptimal scene understanding performance.

In this paper, we take a slightly different viewpoint.
First, inspired by MASt3R~\cite{leroy2024mast3r}, which successfully extracts 3D geometry and pixel correspondences from images, we observe that redundant primitives often share similar geometry and appearance, allowing them to be identified directly from images without relying on complex geometric priors.
Second, we find that per-primitive semantics are not essential for high-performance scene understanding. Instead, a coarse semantic representation, when combined with fine-grained instance information, is sufficient for strong performance, as Gaussian primitives within the same instance exhibit high semantic consistency.

By consolidating this idea, we introduce SpatialSplat, a feed-forward network generating compact yet expressive semantic 3D representations from unposed images. 
As depicted in Fig.~\ref{fig:first-fig} (b), SpatialSplat introduces a dual-field architecture that decomposes the dense semantic feature field into two components: a coarse feature field encoding uncompressed instance-level semantics through minimal Gaussians, and a low-dimensional fine-grained feature field capturing inter-instance relationships.  
Guided by multiple 2D foundation models, our method learns both accurate semantic and instance priors, achieving superior performance while reducing representation parameters.
Additionally, we introduce a Selective Gaussian Mechanism (SGM) to eliminate redundancy in overlapping areas caused by pixelwise representations, along with a novel loss function that jointly optimizes redundancy-aware Gaussians and scene fidelity.
Extensive experiments demonstrate that our method achieves state-of-the-art performance on multiple downstream 3D tasks while using only 40$\%$ of the representation parameters required by the baseline, all without any 3D supervision. 
Our contributions are threefold:
\begin{itemize}
    \item A novel feed-forward 3DGS framework that, to the best of our knowledge, is the first to simultaneously learn semantic and instance priors with guidance from foundation models while maintaining a high-fidelity radiance field.
    
    \item A dual-field semantic representation that significantly reduces storage consumption while enhancing open-vocabulary 3D segmentation performance.  

    \item A mechanism that identifies redundant primitives from images without relying on any geometric priors.
\end{itemize}

%% file: sec/2_related_work.tex
\section{Related Work}
\label{sec:Related}

\begin{figure*}[t]
\vspace{0.3cm}
\centering
 \begin{minipage}{1\linewidth}
    \centering
        \includegraphics[width=1\linewidth]{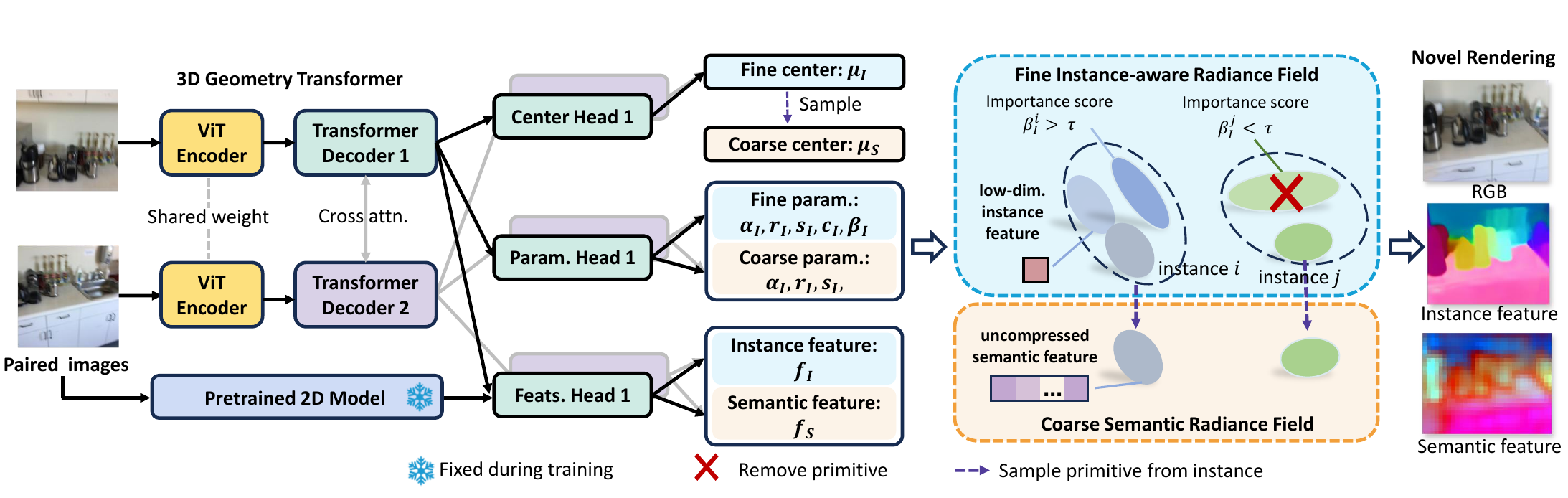}
    \end{minipage}
\vspace{-1em}
\caption{\textbf{Pipeline of SpatialSplat.}
The SpatialSplat processes unposed images along with their intrinsics through a 3D geometry transformer. 
The extracted features from the geometry transformer and the pretrained 2D model are then fed into separate Dense Prediction Transformer (DPT) heads to predict different Gaussian attributes, resulting in a fine-grained instance-aware radiance field $\boldsymbol{\mathcal{F}_I}$ and a coarse semantic feature field $\boldsymbol{\mathcal{F}_S}$.
This enables the synthesis of RGB and feature maps from novel viewpoints. 
}
\label{pipeline}
\vspace{-1.5em}
\end{figure*}

\subsection{Feed-forward 3D Reconstruction}
The emergence of NeRF~\cite{NERF} and 3DGS~\cite{3dgs} has significantly transformed the paradigm of 3D reconstruction from images.
% , eliminating the need for traditional multi-step processes such as feature extraction, matching, and global optimization~\cite{colmap}. 
Despite achieving remarkable results in 3D reconstruction and novel view synthesis, these methods rely on time-consuming per-scene optimization, with even improved techniques~\cite{INGP,K-Planes,Tensorf} still requiring several minutes to hours.
To overcome these limitations, recent approaches~\cite{mvsplat,pixel-splat,pixel-nerf,IBRnet,MuRF,GeoNeRF,Learning,transsplat} have explored using a single feed-forward network to learn reconstruction priors from large datasets, enabling generalizable 3D reconstruction. 
DUSt3R serials~\cite{wang2024dust3r,leroy2024mast3r} further simplifies the reconstruction pipeline, making 3D reconstruction possible from images without relying on camera extrinsics. 
This advancement has inspired a new wave of methods~\cite{ye2024noposplat,smart2024splatt3r,instant-splat} capable of achieving dense reconstruction and view synthesis without camera extrinsics.

\subsection{Feature Field  Distillation}
Early approaches of feature field  distillation were largely based on NeRF~\cite{NERF}. 
For example, methods~\cite{Panoptic-Lifting,SEMANTIC-NERF} successfully embedded 2D label data into 3D space, achieving sharp and precise 3D segmentation. 
Later approaches~\cite{NeRF-SOS,DFF,lerf,Neural-eature-Fusion-Fields,F3RM} incorporated more complex pixel-aligned feature vectors from technologies like LSeg~\cite{lseg} or DINO~\cite{DINO} into NeRF frameworks, enabling open-vocabulary 3D segmentation.
With the rise of 3DGS~\cite{3dgs}, methods such as~\cite{langsplat, Feature3dgs, egolifter, Feature-platting} have extended this concept by embedding language features into Gaussian attributes to enhance semantic understanding.
Large Spatial Model (LSM)~\cite{lsm} was the first to achieve a generalizable semantic radiance field based on feed-forward 3DGS.
However, due to the explicit nature of 3DGS representations, attaching feature vectors to each primitive incurs high memory costs. 
To mitigate this, existing methods often compress semantic features into lower-dimensional representations, inevitably resulting in information loss.

In contrast, our approach decomposes semantics into coarse instance-level semantics and fine-grained instance information, achieving superior performance with significantly lower storage requirements.

\subsection{Compact 3D Gaussian}
Our work is also related to compact 3D Gaussian representation. 
% While 3DGS offers faster rendering speeds compared to NeRF-based methods due to its explicit representation, it requires more storage.
Recent works have attempted to compress 3DGS by exploring the spatial relationships among Gaussians. 
Scaffold-GS~\cite{ScanNeRF} introduces anchor-centered features to reduce parameter counts, while HAC~\cite{HAC} builds on this by incorporating a hash grid to better organize scene primitives. 
Methods~\cite{LightGaussian,Compact-3dgs} utilize techniques like distillation and Gaussian pruning to minimize storage requirements.

Notably, our focus is on addressing the additional representational redundancy introduced by the feed-forward pipeline (i.e., overleaping Gaussians) rather than the inherent redundancy of 3DGS itself. 
This issue is prevalent across existing generalizable 3DGS methods, and the aforementioned compression techniques are not suitable for these frameworks.

%% file: sec/3_method.tex
\section{Method}
\label{sec:Method}
\vspace{2pt}\noindent\textbf{Overview.}
As illustrated in Fig.\ref{pipeline}, SpatialSplat leverages a standard Vision Transformer (ViT)~\cite{ViT} backbone with a few Dense Prediction Transformer (DPT)~\cite{dpt} heads.
Given $V$ unposed images at resolution $H \times W$ with their intrinsics $\{ \boldsymbol{I}^v , \boldsymbol{K}^v \}^V_{v=1}$, SpatialSplat learn a mapping $f_{\theta}$ parameterd by $ \theta $ that transforms  the input images to an 
 fine-grained instance-aware radiance field $\boldsymbol{\mathcal{F}_I}$ and a coarse semantic feature field $\boldsymbol{\mathcal{F}_S}$, which can be formulated as:
\begin{equation}
    \begin{split}
        & f_{\theta} : 
        \{ \boldsymbol{I}^v , \boldsymbol{K}^v \}^V_{v=1} 
         \rightarrow 
        \boldsymbol{\mathcal{F}_I} + \boldsymbol{\mathcal{F}_S},\\
       & \boldsymbol{\mathcal{F}_I} = \{\cup( \boldsymbol{f}_I^i,  \boldsymbol{\beta}^i, \boldsymbol{\mu}^i,\boldsymbol{\alpha}^i,\boldsymbol{r}^i,\boldsymbol{s}^i,\boldsymbol{c}^i)\}_{i=1,...,V \times H\times W}, \\
       &\boldsymbol{\mathcal{F}_S} = \{\cup(\boldsymbol{f}_S^j,\boldsymbol{\mu}^j,\boldsymbol{\alpha}^j,\boldsymbol{r}^j,\boldsymbol{s}^j,\boldsymbol{c}^j)\}_{j=1,...,V \times \frac{H}{S}\times \frac{W}{S}}.
    \end{split}
\end{equation}  
$\boldsymbol{\beta} \in \mathbb{R}$ is the importance score, which will be explained in Section~\ref{method:Pixel-free}.
$\boldsymbol{f}_I \in \mathbb{R}^N$ is the learnable instance feature and $\boldsymbol{f}_S \in \mathbb{R}^M$ is the learnable semantic feature, as detailed in Section~\ref{method:Semantic_Radiance_Field}.
% All of these will be detailed in Section 1.
The remains represent the vanilla Gaussian parameters~\cite{3dgs}, including the center position $\boldsymbol{\mu} \in \mathbb{R}^3$, opacity $\boldsymbol{\alpha} \in \mathbb{R}$, rotation factor in quaternion $\boldsymbol{r} \in \mathbb{R}^4$, scale $\boldsymbol{s} \in \mathbb{R}^3$,
and spherical harmonics (SH) $\boldsymbol{c} \in \mathbb{R}^k$ with $k$ degrees of freedom.
% Primitives in $\boldsymbol{\mathcal{F_I}}$ with importance scores below a threshold are discarded, while the remaining ones are rendered into novel views and instance feature maps with a resolution of $H \times W$.
% To better capture scene semantics, features from pre-trained 2D vision models are injected into  $\boldsymbol{\mathcal{F_S}}$, enabling the rendering of a coarse semantic feature map with high accuracy.
In the following sections, we provide a detailed explanation of each component of our method.

\subsection{3D Geometry Prediction}
\label{method:Geometry-Prediction}
Both the encoder and decoder in our geometric prediction module are built on pure ViT structures, requiring no geometric priors as in previous methods~\cite{pixel-splat,mvsplat}.
The input image is patchified and flattened into image sequences, which along with the camera intrinsics processed by a linear layer, are fed into the encoder.
The encoder weights are shared across different input views.
The features from encoder are then passed to a ViT-based decoder, where cross-attention is applied to better capture spatial relationships and aggregate information across views.
Finally, the features from different views are separately fed into a series of DPT heads to predict point maps and other Gaussian parameters.
The point maps serve as Gaussians centers.
Since the input images are unposed, the Gaussians centers suffer from scale ambiguity.
Some methods~\cite{lsm,smart2024splatt3r} rely on depth supervision to obtain per-scene scale, but this introduces additional errors due to imperfect depth.
To overcome this limitation, we follow NoPoSplat’s~\cite{ye2024noposplat} approach by injecting camera intrinsics to resolve scale ambiguity.
% in a canonical space.
% Specifically, we designate one input view’s camera coordinate system as the canonical space and train the network to predict all Gaussian centers within this space.
% The camera intrinsics are injected here to resolve scale ambiguity.
% , enabling direct rendering of target views without the per-scene scale estimation.
Experiments show that SpatialSplat effectively learns 3D priors from sparse unposed images without depth supervision, even while jointly learning multiple parameters and features.
% This enable our method effectly learning 3D information from only multi views, and further eliminates the cumulative errors introduced by align views by addition 3D information.
% required in DUSt3R~\cite{wang2024dust3r}.
% This approach further eliminates the cumulative errors introduced by scale estimation, leading to improved performance.

\subsection{Selective Gaussian Mechanism}
\label{method:Pixel-free}
Previous methods~\cite{smart2024splatt3r,ye2024noposplat,lsm, pixel-splat,mvsplat} directly use the pixel-wise pointmap as Gaussian centers which introduce significant redundancy.
To address this, we propose a selective Gaussian mechanism that assigns each primitive an importance score to quantify its necessity for the scene representation.
Primitives with importance scores below a certain threshold are considered redundant and discarded.
% Directly supervising the learning of importance score is challenging, as defining a function to quantify the significance of each primitive is not straightforward~~\xnote{tell more, ``straightforward'' is not enough for a new reader.}.
To effectively supervise the learning of importance scores, we analyze Gaussian properties and the rendering pipeline, designing a simple yet effective strategy.
Specifically, a primitive's influence can be reduced by decreasing its opacity or size~\cite{3dgs}.
However, we observe that even primitives with near-zero size still contribute to the rendering by occupying a pixel.
Therefore, we multiply the importance score to the opacity of each primitive and modify the alpha blending formulation as follows:
\begin{equation}
\label{alpha-blending}
    \boldsymbol{c} = \sum_{i=1}^{n}\boldsymbol{c}_i\boldsymbol{\alpha}_i\boldsymbol{\beta}_i\prod_{j=1}^{i-1}(1-\boldsymbol{\alpha}_j\boldsymbol{\beta}_j),
\end{equation}  
where $\boldsymbol{c}$ is the final intensities calculated by blending $n$ ordered Gaussians overlapping the pixels, and $\boldsymbol{\beta}_i$ is the importance score processed through a sigmoid activation function.
% ~\xnote{Is this $\boldsymbol{\beta}_i$ the same as the previous $\boldsymbol{\beta}_i$ a bit earlier in this paragraph? If they are the same, do not use $\boldsymbol{\beta}_i$ before, because we need to explain it when it first appears. }
This ensures that primitives with low importance have minimal impact on the final intensities.
Removing important primitives leads to a significant increase in photometric loss, whereas discarding redundant ones has little effect.
Therefore, we optimize $\boldsymbol{\beta}_i$ through photometric loss minimization.
Following methods~\cite{lsm,ye2024noposplat}, we define the photometric loss as a weighted sum of MSE and LPIPS~\cite{LPIPS}:
\begin{equation}
    \mathcal{L}_{P}(\boldsymbol{C},\hat{\boldsymbol{C}}) = \mathcal{L}_{MSE}(\boldsymbol{C},\hat{\boldsymbol{C}}) + \lambda \mathcal{L}_{LPIPS}(\boldsymbol{C},\hat{\boldsymbol{C}}),
\end{equation}
where $\boldsymbol{C}$ and $\hat{\boldsymbol{C}}$ are rasterized and GT pixel intensities, and $\lambda = 0.05$.
Since redundant primitives are directly discarded during inference, we encourage their importance scores to approach zero to minimize the discrepancy between training and inference.
Simultaneously, we aim to reduce the number of primitives used, striving to eliminate redundant ones as much as possible.
To achieve these objectives, we first design the importance score with inspiration from Leaky ReLU~\cite{maas2013leakyrelu}, formulated as follows:
\begin{equation}
    \boldsymbol{\beta_i} = 
    \begin{cases}
    \begin{split}
        & \boldsymbol{\beta}_i\quad &\text{if }\boldsymbol{\beta}_i> \tau\\
        & \boldsymbol{\beta}_i \times 10^{-3} \quad &\text{if }\ \boldsymbol{\beta}_i< \tau
    \end{split}
    \end{cases}.
\end{equation}
This ensures that the importance scores of primitives below the threshold $\tau$ are close to zero while preventing gradient vanishing.
Additionally, we introduce a Binary Cross Entropy (BCE) loss with a regularization term that pushes $\boldsymbol{\beta}_i$ to either 0 or 1 while penalizing excessive use of primitives:
\begin{equation}
\mathcal{L}_{I}(\boldsymbol{S}) = \mathcal{L}_{BCE}(\boldsymbol{S}, \hat{\boldsymbol{S}}) + \frac{1}{\| \boldsymbol{S}\|} \sum_{\boldsymbol{\beta}_i \in \boldsymbol{S}} \boldsymbol{\beta}_i,
\end{equation}
where
\vspace{-0.5em}
\begin{equation*}
\boldsymbol{S} = \{ \boldsymbol{\beta}_i\}_{i=1}^{H \times W}, \quad
\hat{\boldsymbol{S}} = 
    \begin{cases}
    \begin{split}
        & 1 \quad \text{if }\ \boldsymbol{\beta}_i> \tau\\
        & 0 \quad \text{if }\ \boldsymbol{\beta}_i< \tau
    \end{split}
    \end{cases}.
\end{equation*}
As a result, SpatialSplat achieves a compact yet high-fidelity scene representation by selectively retaining only the most essential primitives.

\subsection{Dual-field Architecture}
\label{method:Semantic_Radiance_Field}
While distilling 2D semantic features into dense 3D representations has proven effective for scene understanding, conventional per-primitive compressed feature assignments~\cite{Feature3dgs,langsplat,F3RM} result in information loss and suboptimal performance.
To mitigate this loss without increasing storage costs, we propose a dual-field architecture that decouples semantic representation into: 1) a fine-grained instance-aware radiance field, capturing scene geometry, textures, and instance correspondences, and 2) a coarse, uncompressed semantic feature field that reduces storage requirements while preserving semantic accuracy.
% inspired by~\cite{Feature-platting}. ~\xnote{Do we need to mention this paper? It seems to reduce our contribution because we are also not the first one to use a feed-forward approach.}
% Unlike~\cite{Feature-platting}, which relies on a multi-step process and time-consuming per-scene optimization, we achieve this with a simple feed-forward approach.

\begin{table*}[t]
\small
\vspace{1em}
    \setlength{\abovecaptionskip}{0.3em}
    \centering
    \resizebox{\linewidth}{!}{
    \begin{tabular}{{l}*{9}{c}}
        \toprule
        &  & & \multicolumn{2}{c}{Source View} &  \multicolumn{5}{c}{Target View} \\
        \cmidrule(r){4-5} \cmidrule(r){6-10}
        Methods & Semantic & Forward & mIoU $\uparrow$ & Acc. $\uparrow$ & mIoU $\uparrow$ & Acc. $\uparrow$ & PSNR $\uparrow$ & SSIM $\uparrow$ & LPIPS $\downarrow$ \\
        \midrule
        L-Seg~\cite{lseg} &\ding{52} &\ding{55}  & 0.5541 & 0.7824 & 0.5558 & 0.7856 & N/A & N/A & N/A \\
        NeRF-DFF~\cite{nerfdff} &\ding{52} &\ding{55} & 0.5381 & 0.7724 & 0.5137 & 0.7582 & 22.49 & 0.7650 & 0.2829 \\
        Feature-3DGS~\cite{Feature3dgs} &\ding{52} &\ding{55} & 0.4992 & 0.7362 & 0.3223 & 0.5674 & 17.96 & 0.5812 & 0.4894 \\
        \midrule
        NoPoSplat~\cite{ye2024noposplat} &\ding{55} &\ding{52} & N/A & N/A & N/A & N/A & 25.70 & 0.8159 & 0.1875 \\ 
        \midrule
        LSM~\cite{lsm}  &\ding{52} &\ding{52} & 0.5141 & 0.7719 & 0.5104 & 0.7662 & 24.12 & 0.7961 & 0.2526 \\
        SpatialSplat-Lite &\ding{52} &\ding{52} & 0.5272 & 0.7801 & 0.5265 & 0.7693 & 25.45 & 0.8032 & 0.2039\\
        \rowcolor{myblue} SpatialSplat&\ding{52} &\ding{52} & 0.5593 & 0.7814 & 0.5587 & 0.7924 & 25.46 & 0.8045 & 0.2046\\
         \bottomrule
         % \multicolumn{10}{l}{\small {$\ast$} denotes we replace the pretrained 2D model LSeg with CLIP ViT-B/16.}\\
    \end{tabular}
    }
\caption{\textbf{Quantitative Comparison in 3D Tasks on Scannet dataset}. Our method outperforms both the latest SOTA semantic-aware feed-forward approach and per-scene optimization methods. ``-Lite'': replacing the pretrained 2D model LSeg with CLIP ViT-B/16.}
\vspace{-0.3cm}
\label{table:Quantitative}
\end{table*}

\vspace{2pt}\noindent\textbf{Fine-grained Instance-Aware Radiance Field.}
\label{method:instance-field}
Our instance-aware radiance field builds upon Vanilla 3DGS, initializing the redundancy-aware Gaussians augmented with learnable instance features $\boldsymbol{f}_I$.
We train the network using contrastive learning~\cite{Contrastive-Lift} to pull rendered instance features of the same instance closer while push those of different instances further apart.
To achieve this, we lift multi-view 2D segmentation priors into 3D to help SpatialSplat learn inter-instance relationships.
Given a target image $\boldsymbol{I}$, we first use Segment Anything Model (SAM)~\cite{sam} to extract a set of masks 
$\mathcal{M} = \{ \boldsymbol{M^k} \mid k = 1,2,...,m \}$.
The instance feature map rendered at target image view with Eq.~\ref{alpha-blending} is denoted as $\boldsymbol{F}_I$.
The contrastive loss is then defined as:
\begin{equation}
% \small
\label{loss:constrast}
    \begin{split}
        \mathcal{L}_{C}(\boldsymbol{F}_I,\mathcal{M}) & = -log \frac{1}{\| \boldsymbol{I} \| ^2} \sum_{u \in \boldsymbol{I}} \sum_{u^{'} \in \boldsymbol{I}^+} sim(\boldsymbol{F}_I(u),\boldsymbol{F}_I(u^{'}) )\\
        & + \sum_{u^{'} \notin \boldsymbol{I}^+} (1-sim(\boldsymbol{F}_I(u),\boldsymbol{F}_I(u^{'}) )),
    \end{split}
\end{equation}
where $\boldsymbol{I}^+$ is the set of pixels that belong to the same instance mask, 
$\boldsymbol{F}_I(u)$ is the feature vector of the $\boldsymbol{F}_I$ at coordinate $u$ and $sim(.)$ is the cosine similarity.
With this weak supervision, SpatialSplat effectively clusters radiance primitives by instance affiliation.
Notably, the loss complexity in Eq.~\ref{loss:constrast} scales quadratically with the number of pixels, making training infeasible. To address this, we employ a trick (detailed in the appendix) that enables efficient loss estimation with linear complexity.
% The complexity of the above formulation is proportional to the square of the number of pixels. 
% Therefore, we use a randomly sampled set of pixels 
% $\mathcal{U} \in \boldsymbol{I}$ to replace 
% $\boldsymbol{I}$ due to GPU memory limitations.
% \xnote{Maybe we can move these to the implementation section. Does this belong to technical points, or are these small tricks for implementation?}
% ~\xnote{I remember we claimed to have eliminated 3D supervision before. It is confusing.}
% As a result, the network only needs to learn semantic information at the instance level rather than for each individual primitive.

\vspace{2pt}\noindent\textbf{Coarse Semantic Feature Field.}
Since primitives within the same instance share semantic consistency, SpatialSplat learns instance-level semantics using only a subset of primitives, avoiding the redundancy of per-primitive assignments.
Manually selecting primitives for each instance, however, disrupts the network's cohesion and simplicity while increasing additional computational costs.
To address this, we uniformly downsample the Gaussian centers from $\boldsymbol{\mathcal{F}_I}$ to form a coarse feature field, where each selected primitive is assigned a semantic feature $\boldsymbol{f}_S$.
To enhance semantic learning, we inject image features from a pretrained 2D model into feature heads, as shown in Fig.~\ref{pipeline}.
A 2D semantic feature map  $\boldsymbol{F}_S$ is rendered using alpha blending.
% as follows:
% \begin{equation}
%     \boldsymbol{s} = \sum_{i=1}^{n} 
%     {^{S}\boldsymbol{f}_i} \boldsymbol{\alpha}_i
%     \prod_{j=1}^{i-1}(1-\boldsymbol{\alpha}_j
%     ),
% \end{equation}
% $\boldsymbol{s}$ represents the final semantic feature for a pixel, and 
% $^{S}\boldsymbol{f}_i$
% is the feature of the $i$-th Gaussian that overlaps with the pixel. 
We minimize the loss between the rendered feature map at a novel view and the feature map $\hat{\boldsymbol{F}}_S$ of the ground truth image extracted from the pretrained 2D model:
\begin{equation}
    \mathcal{L}_{S}(\boldsymbol{F}_S, \hat{\boldsymbol{F}}_S) = MSE(\boldsymbol{F}_S, \hat{\boldsymbol{F}}_S)
\end{equation}

During open-vocabulary querying, we first cluster primitives in $\boldsymbol{\mathcal{F}_I}$  by instance feature similarity, assigning each instance the average feature of its clustered primitives as its label.
For primitives in $\boldsymbol{\mathcal{F}_S}$, their labels correspond to the instance features before sampling, and their semantic features are assigned to the instance with the highest cosine similarity.
If multiple primitives contribute semantic features to the same instance, we take their average to ensure consistency.
This approach enables the network to acquire sharper and clearer semantics compared to the privious methods.
An additional advantage of dual-field architecture is that, unlike dense semantic supervision from LSeg~\cite{lseg}, it allows the use of a much lighter pretrained model (e.g., CLIP ViT-B/16~\cite{CLIP}), significantly improving inference speed.

\subsection{Training Objective}
Following previous methods~\cite{lsm}, we perform end-to-end training to optimize our model with the following objective:
\begin{equation}
    \begin{split}
        \mathcal{L} & = \mathcal{L}_{P}(\boldsymbol{C},\hat{\boldsymbol{C}}) + \lambda_1 \mathcal{L}_{I}(\boldsymbol{S}) \\
        &+ \lambda_2\mathcal{L}_{C}(\boldsymbol{F}_I,\mathcal{M}) + \lambda_3 \mathcal{L}_{S}(\boldsymbol{F}_S, \hat{\boldsymbol{F}}_S).
    \end{split}
\end{equation}
The parameters $\lambda_1$, $\lambda_2$, $\lambda_3$ are set to 0.01, 0.2, and 1.0, respectively.
To save training time, the instance masks $\mathcal{M}$ are generated by SAM~\cite{sam} prior to training, while the semantic feature map $\hat{\boldsymbol{F}_S}$ is predicted during training.
% The input image extrinsics are referenced to the first image, with the baseline between extrinsics scaled to 1. Since SpatialSplat predicts Gaussian primitives in canonical space, it eliminates the need for ground truth depth for scale estimation during inference. 
% Moreover, it does not require the test-time pose alignment used in NoPoSplat~\cite{ye2024noposplat} inference.\xnote{It is sudden to bring up NoPoSplat. Is it important to emphasize this point?}
\begin{figure*}[t]
\vspace{-0.5em}
\centering
\begin{minipage}{1\linewidth}
    \centering
        \includegraphics[width=1\linewidth]{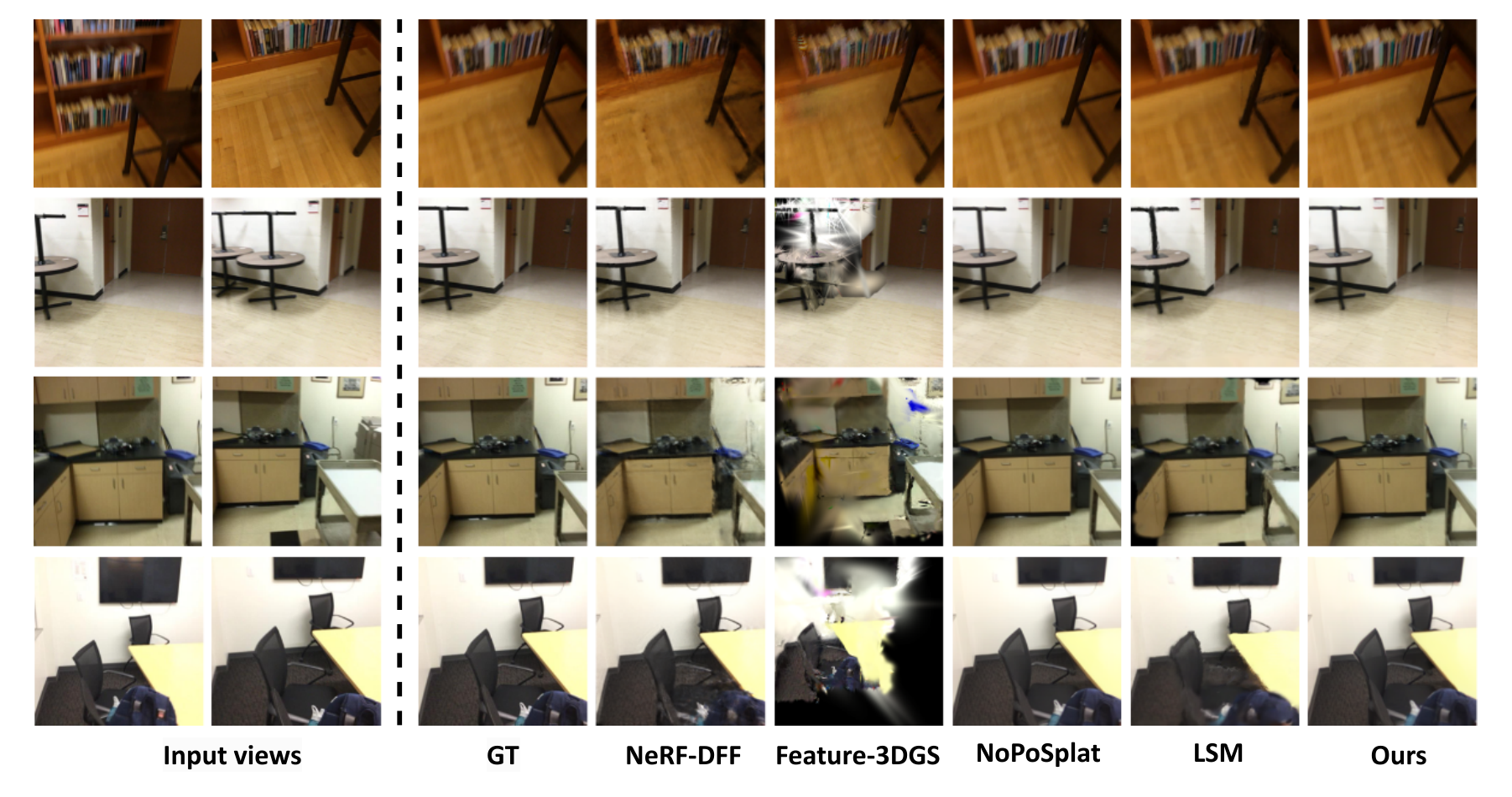}
\end{minipage}
\vspace{-1.5em}
\caption{\textbf{Qualitative comparison in NVS.} SpatialSplat can synthesize realistic novel views. In challenging cases where LSM fails, such as the table legs in the first two rows and the corners in the last two rows, our method achieves significantly better results.}
\label{fig:nvs}
% \vspace{-0.5em}
\end{figure*}

\begin{table*}[t]
\small
% \vspace{1em}
    \setlength{\abovecaptionskip}{0.3em}
    \centering
    \resizebox{\linewidth}{!}{
    \begin{tabular}{{l}*{10}{c}}
        \toprule
         & \multicolumn{5}{c}{LSM} &  \multicolumn{5}{c}{SpatialSplat (Ours)} \\
        \cmidrule(r){2-6} \cmidrule(r){7-11}
        Scene & mIoU $\uparrow$ & Acc. $\uparrow $ &  PSNR $\uparrow$ & SSIM $\uparrow$ & LPIPS $\downarrow$ & mIoU $\uparrow$ & Acc. $\uparrow $ &  PSNR $\uparrow$ & SSIM $\uparrow$ & LPIPS $\downarrow$\\
        \midrule
        room0 & 0.5024 & 0.7423 & 17.61 & 0.5247 & 0.3228 & \textbf{0.5131} & \textbf{0.7934} & \textbf{22.09} & \textbf{0.6760} &  \textbf{0.1794}\\
        room1 & 0.4661 & 0.7349 & 14.34 & 0.5913 & 0.3174 & \textbf{0.4691} & \textbf{0.7628} & \textbf{20.62} & \textbf{0.6774} & \textbf{0.2265}\\
        office3 & 0.5484 & 0.7627 & 21.28 & 0.7567 & 0.2084 & \textbf{0.5582} & \textbf{0.7997} & \textbf{22.52} & \textbf{0.7913} & \textbf{0.1436} \\
        office4 & \textbf{0.5065} & \textbf{0.7723} & 19.51 & 0.6962 & 0.2779 & 0.5016 & 0.7695 & \textbf{21.13} & \textbf{0.7271} & \textbf{0.2072}\\
         \bottomrule
    \end{tabular}
    }
\caption{\textbf{Out-of-distribution (OOD) comparison on Replica dataset.} SpatialSplat generalizes well on OOD data.}
\vspace{-0.3em}
\label{table:Out-of-distribution}
\end{table*}

%% file: sec/4_exp.tex
\section{Experiment}
\label{sec:Experiment}

\begin{figure*}[t]
% \vspace{0.3cm}
\centering
 \begin{minipage}{1\linewidth}
    \centering
        \includegraphics[width=1\linewidth]{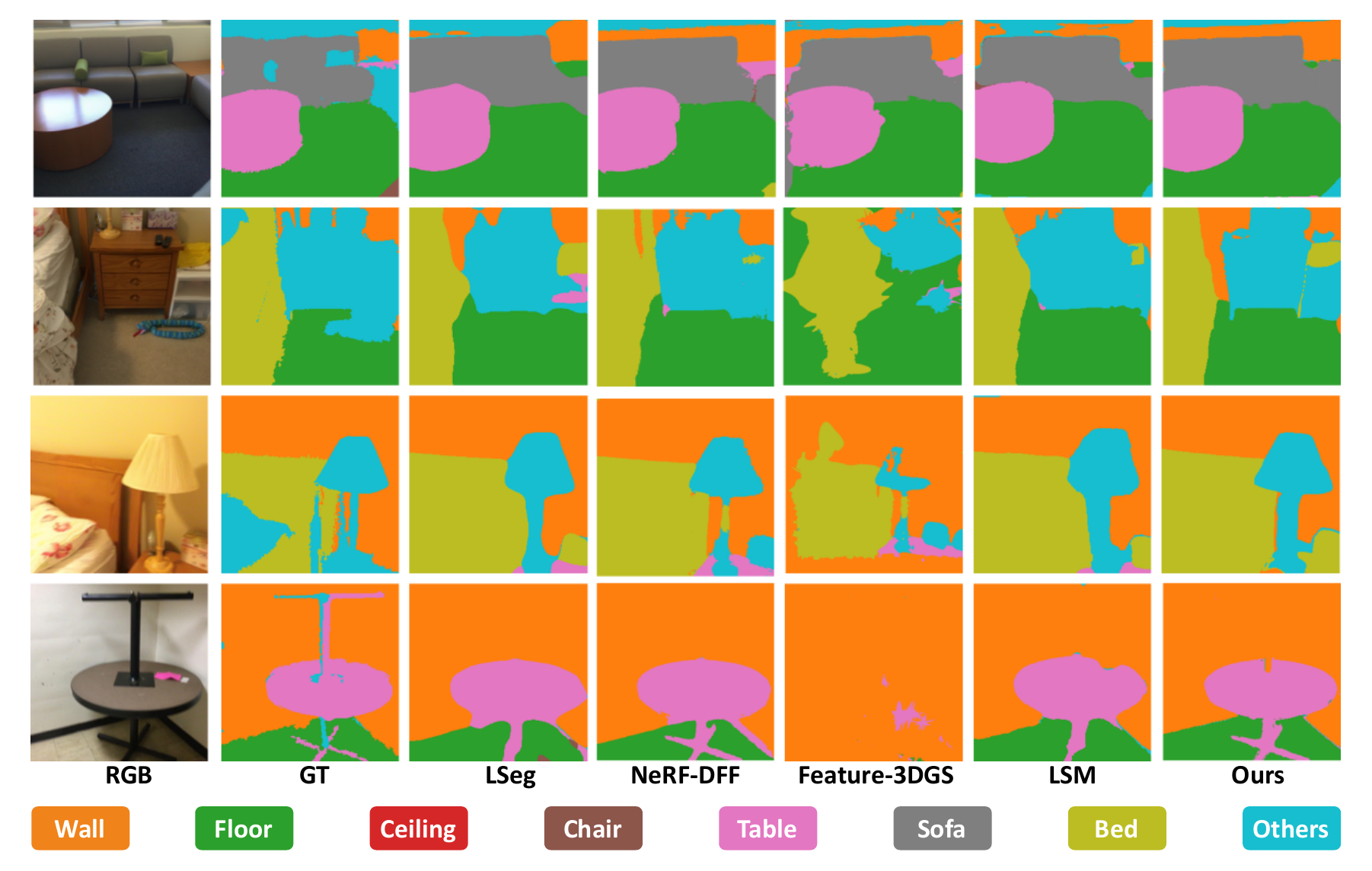}
    \end{minipage}
\vspace{-1em}
\caption{\textbf{Qualitative comparison in OVS.} SpatialSplat achieves sharper and more precise segmentation results compared to previous methods. Notably, our method excels in challenging details, such as distinguishing table legs from cabinet legs (the second the fourth rows), benefiting from our dual-field architecture that captures detailed instance information and uncompressed semantics.}
\label{fig:ovs}
\vspace{-0.5em}
\end{figure*}

\begin{figure}[t]
% \vspace{-1em}
\centering
 \begin{minipage}{1\linewidth}
    \centering
        \includegraphics[width=1\linewidth]{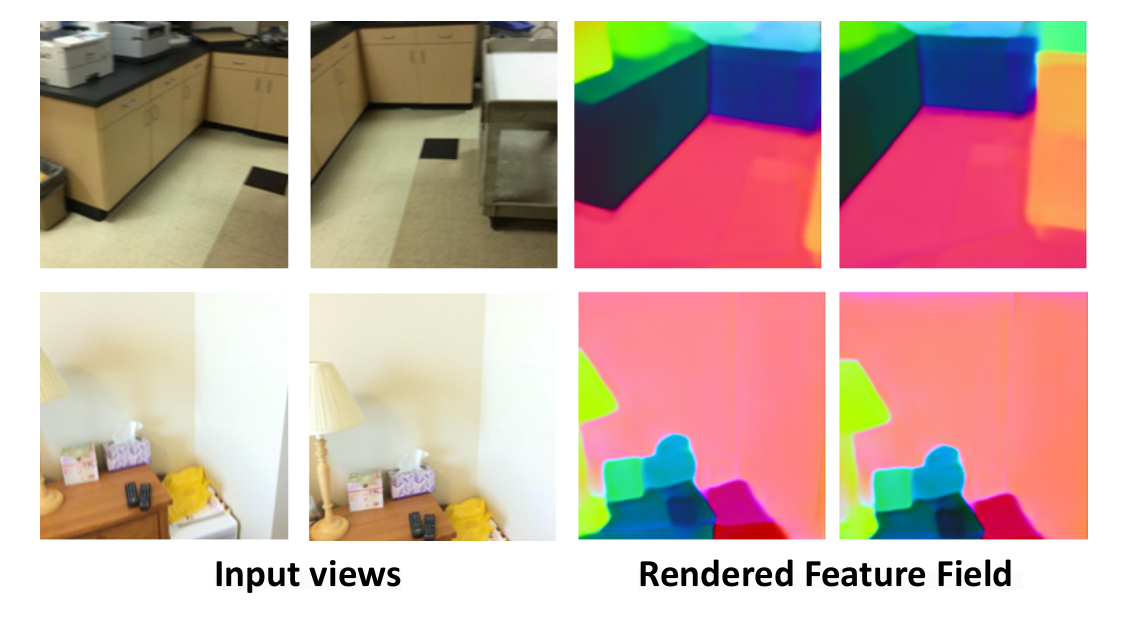}
    \end{minipage}
    \vspace{-1em}
\caption{\textbf{The rendered instance features.} SpatialSplat predicts clear and consistent instance features across different views.}
\label{fig:vis_instance}
\vspace{-1em}
\end{figure}

\begin{figure}[t]
% \vspace{0.3cm}
\centering
 \begin{minipage}{1\linewidth}
    \centering
        \includegraphics[width=1\linewidth]{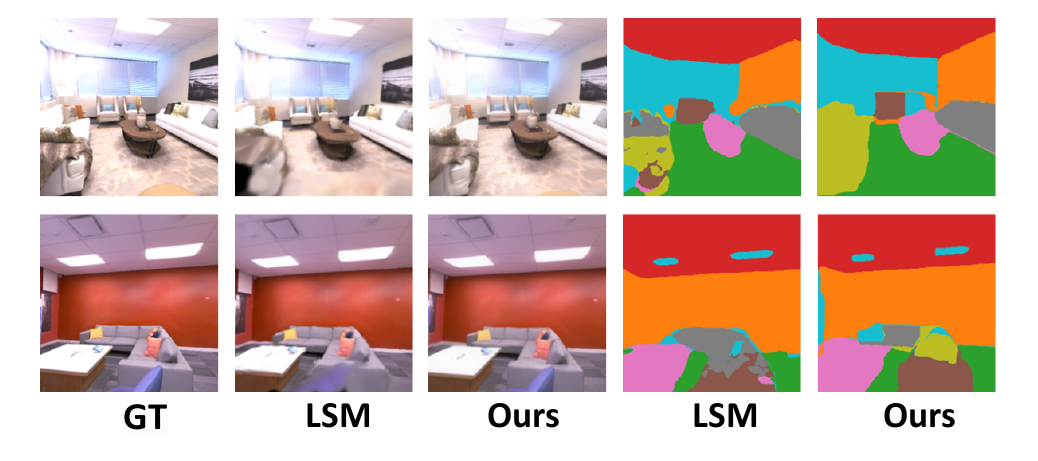}
    \end{minipage}
\vspace{-1.5em}
\caption{\textbf{Qualitative results of cross-dataset generalization.} Zoom out for clearer visualization.}
\label{fig:cross-data}
% \vspace{-2em}
\end{figure}

\subsection{Experimental Setup}
\vspace{2pt}\noindent\textbf{Datasets.}
Following LSM\cite{lsm}, we primarily train our model on a combination of ScanNet++\cite{yeshwanthliu2023scannetpp} and ScanNet~\cite{dai2017scannet}.
We filter out bad scenes and those with incomplete extrinsic parameters, resulting in a training dataset of approximately 1,500 scenes.
For evaluation, we follow LSM and select 40 unseen scenes from ScanNet to assess our model's performance.
We test our method's cross-dataset generalization on the synthesized Replica Dataset~\cite{replica19arxiv}.

\vspace{2pt}\noindent\textbf{Evaluation Metrics.}
We evaluate our method on two downstream tasks: novel view synthesis (NVS) and 3D open-vocabulary segmentation (OVS).
For NVS, we adopt the standard metrics: PSNR, SSIM, and LPIPS~\cite{LPIPS}.
For OVS, we evaluate performance using class-wise intersection over union (mIoU) and average pixel accuracy (mAcc).

\vspace{2pt}\noindent\textbf{Baselines.}
We primarily compare our method with LSM~\cite{lsm}, the latest state-of-the-art (SOTA) approach for generalizable semantic 3D reconstruction.
Additionally, we include the following baselines:
1): Feature-3DGS~\cite{Feature3dgs} and NeRF-DFF~\cite{nerfdff}, pre-scene optimization methods for semantic 3D reconstruction based on 3DGS~\cite{3dgs} and NeRF~\cite{NERF}, respectively.
2): NoPoSplat~\cite{ye2024noposplat}, a latest SOTA method for generalizable novel view synthesis.
3): LSeg~\cite{lseg}, a 2D open-vocabulary segmenter used for feature distillation in all compared semantic-aware methods.
Unless otherwise specified, our method also uses image features from LSeg for supervision.

\vspace{2pt}\noindent\textbf{Implementation details.}
% We implement SpatialSplat using PyTorch. 
For the 3D geometry prediction module, we use ViT-Large with a patch size of 16 as the encoder and ViT-Base as the decoder, both initialized with pre-trained weights from MASt3R~\cite{leroy2024mast3r}, while the remaining modules are randomly initialized.
We modify the 3DGS CUDA renderer to support instance and semantic feature map rendering. 
The importance score threshold $\tau$ is set to 0.5, the downsampling ratio to 8, the instance feature dimension $N$ to 8, and the semantic feature dimension $M$ to 512.
For a fair comparison, we train the model at a resolution of 256 × 256, consistent with our baseline.
% All models are trained on eight A100-40 GPUs using the Adam optimizer with the learning rate of $10^{-4}$.
% More details can be found in the appendix.
% , with an initial learning rate of $2 \times 10^{-5}$ for the backbone and $2 \times 10^{-4}$ for other parameters.

\subsection{Results and Analysis}
\vspace{2pt}\noindent\textbf{Results of Novel View Synthesis.}
Following previous methods~\cite{ye2024noposplat,pixel-splat}, we select two images from each test scene as input view for the generalizable 3DGS methods, and choose four additional images located between these two as target views to evaluate the performance of different methods in Novel View Synthesis. 
Notably, Feature-3DGS~\cite{Feature3dgs} and NeRF-DFF~\cite{nerfdff} relies on dense view inputs and performs poorly with only two input images. 
To account for this, we use five out of the six available images (the two input images and four target views) for training and evaluate on the remaining image.
As shown in Tab.~\ref{table:Quantitative}, SpatialSplat significantly outperforms latest SOTA method LSM. 
We observe that LSM struggles in certain scenes due to its reliance on accurate depth for aligning views during training as shown in Fig.~\ref{fig:nvs}. 
% However, despite the dataset providing true depth values, noise still exists, making it challenging to generate consistent views in some scenes. 
% In contrast, our method directly predicts Gaussians in a canonical space, eliminating alignment errors and delivering superior performance. 
Although NeRF-DFF uses more images for training, our method still outperforms it and even demonstrates competitive performance with very recent SOTA methods designed specifically for novel view synthesis. 
This highlights that our method efficiently learns spatial priors from large-scale 2D data.

\vspace{2pt}\noindent\textbf{Results of Open-vocabulary 3D Segmentation.}
Following LSM's approach, we map category labels from the Scannet dataset into common categories: Wall, Floor, Ceiling, Chair, Table, Bed, Sofa, Others. 
As shown in Tab.~\ref{table:Quantitative}, SpatialSplat outperforms all compared methods, even surpassing L-Seg, which provides semantic feature supervision for other compared methods. 
As illustrated in Fig.~\ref{fig:ovs}, SpatialSplat produces sharp and clear 3D semantic segmentation results. 
This is achieved by lifting both 2D instance and uncompressed semantic into 3D space, enabling more efficient semantic learning while mitigating the blurred contours and information loss associated with per-primitive compressed semantics.
The visualizations in Fig.~\ref{fig:vis_instance} show that SpatialSplat effectively learns a highly consistent instance prior with only weak supervision from 2D masks. 
% Primitives within the same instance exhibit strong feature coherence, while those from different instances remain distinctly separated. 
% This serves as the foundation for the sharp and well-defined semantic feature maps presented in Fig.~\snote{to be done}.

\begin{figure}[t]
\vspace{0.3cm}
\centering
 \begin{minipage}{1\linewidth}
    \centering
        \includegraphics[width=1\linewidth]{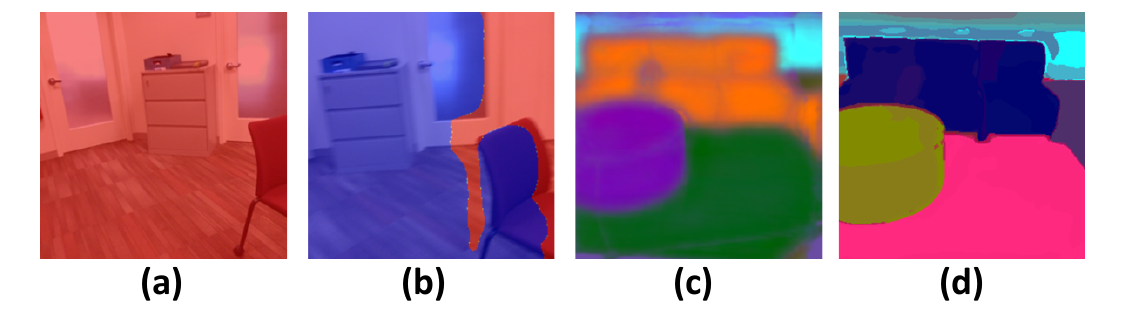}
    \end{minipage}
\vspace{-1em}
\caption{\textbf{Qualitative results of ablations.} (a) and (b) Qualitative results of importance score prediction, with red color indicating an importance score of 1 and blue indicating 0.
(c) The rendered semantic feature without the dual-field.
(d) The rendered semantic feature with the dual-field, which appears clearer and sharper.}
\label{fig:ablation}
\vspace{-1em}
\end{figure}

\begin{table}[t]
\small
\vspace{1em}
    \setlength{\abovecaptionskip}{0.3em}
    \small
    \centering
    \resizebox{\linewidth}{!}{
    \begin{tabular}{{l}{c}{c}{c}}
        \toprule
         % &  \multicolumn{2}{c}{Efficiency} \\
        % \cline{2-3}
        Method & Latency$\downarrow$ & Gaussian Size $\downarrow$ & Num. $\downarrow$ \\
        \midrule
        Feature-3DGS~\cite{Feature3dgs} & 1069 s & 229.65 MB & 476K \\
        NoPoSplat~\cite{ye2024noposplat} & 0.044 s & 30.93 MB & 131K\\
        LSM~\cite{lsm} & 0.104 s &  63.00 MB & 131K\\
        SpatialSplat-Lite & \textbf{0.058 s } & 25.58 MB & 87.7K \\
        \rowcolor{myblue} SpatialSplat & 0.071 s & \textbf{25.58 MB} & \textbf{87.7K} \\
         \bottomrule
    \end{tabular}
    }
\caption{\textbf{Model efficiency comparison.} ``-Lite'': the model replaces LSeg with CLIP ViT-B/16.}
\label{table:Model-Efficiency}
\vspace{-1.2em}
\end{table}

\vspace{2pt}\noindent\textbf{Results of Cross-Dataset Generalization.}
As shown in Tab.~\ref{table:Out-of-distribution}, our method significantly outperforms LSM on the unseen Replica dataset in both novel view synthesis and open-vocabulary segmentation. 
Notably, since Replica is a synthetic dataset with a different data modality from our training set, this underscores the strong generalization ability of our approach. 
The visualization in Fig.~\ref{fig:cross-data} further demonstrates that LSM struggles with coarse modeling at scene edges, leading to artifacts in synthesized views, whereas our method generates photorealistic renderings.

\vspace{2pt}\noindent\textbf{Analysis of Model Efficiency.}
We compare the model efficiency of different 3DGS-based methods. 
As shown in Tab.~\ref{table:Model-Efficiency}, our method is faster than LSM thanks to its streamlined architecture, requiring only 40$\%$ of the storage size and 65$\%$ of primitive number.
% while achieving superior performance on downstream 3D tasks.
Remarkably, our approach requires even fewer Gaussians parameters than NoPoSplat, which lacks semantic awareness. 
This demonstrates the effectiveness of our proposed dual-field architecture and selective Gaussian mechanism, which capture accurate semantics while eliminating redundant Gaussians.
Furthermore, as our method does not rely on dense semantic supervision, we leverage a lightweight pretrained 2D model, significantly accelerating inference speed.
% In contrast, per-scene optimization methods focus on fitting each individual view, which leads to an excessive number of disorganized primitives and ultimately reduces efficiency.

% \begin{table*}[t]
% \vspace{1em}
%     \setlength{\abovecaptionskip}{0em}
%     \centering
%     \caption{\textbf{Combine With Existing Methods}}
%     \label{table:Combine}
%     % \resizebox{\linewidth}{!}{
%     \begin{tabular}{*{5}{c}}
%         \toprule
%         Training Data &  Method & PSNR \uparrow & SSIM \uparrow & LPIPS \downarrow \\
%         \midrule
%         Re10K & NoPoSplat & - & - & -\\
%         Re10K & NoPoSplat + Pixel-free & - & - & -\\
%         Scannet++ & Splat3R & - & - & - \\
%         Scannet++ & Splat3R + Pixel-free& - & - & - \\
%          \bottomrule
%     \end{tabular}
%     % }
% \vspace{-2em}
% \end{table*}

\subsection{Ablations and Analysis.}
 We perform ablations to answer the following questions:
(1) Are the primitives removed by our selective Gaussian mechanism truly redundant?
(2) Does the improved scene understanding stem from the proposed dual-field architecture?
(3) As the number of views increases, how does redundancy grow? Can our method effectively alleviate it?

% \begin{table*}[t]
% \vspace{1em}
%     \setlength{\abovecaptionskip}{0em}
%     \centering
%     \caption{\textbf{Ablation Study on Our Design Choices}}
%     \label{table:Ablations}
%     % \resizebox{\linewidth}{!}{
%     \begin{tabular}{*{7}{c}}
%         \toprule
%        Model &  mIoU\uparrow & Acc.\uparrow & PSNR\uparrow  & SSIM \uparrow & LPIPS \downarrow & Param. \downarrow\\
%         \midrule
%        Ours & 0.5577 & 0.7920 &  25.4718 & 0.8046 & 0.2039 & 45k\\
%        w/o. Pixel-free & 0.5569 & 0.7972 & 25.5550 & 0.8041 & 0.1993& 65k\\
%        w/o. Decouple & 0.5027 & 0.7786 & 25.4693 & 0.8073 & 0.2021 & 45k\\
%        Multi-view & - & - & 26.5708 & 0.8137 & 0.1832 & \\
%          \bottomrule
%     \end{tabular}
%     % }
% \vspace{-2em}
% \end{table*}

\begin{table}[t]
\small
% \vspace{1em}
    \setlength{\abovecaptionskip}{0.3em}
    \centering
    \resizebox{\linewidth}{!}{
    \begin{tabular}{{l}{c}{c}{c}{c}}
        \toprule
       Model &  mIoU$\uparrow$ & Acc.$\uparrow$ & PSNR$\uparrow$ & Num. $\downarrow$\\
        \midrule
       No SGM & 0.5569 & 0.7972 & 25.55 & 131k\\
       No dual & 0.5027 & 0.7786 & 25.46 &  87.6k\\
       3-view & 0.5125 & 0.7793 & 24.92 &  97.5k \\
       \rowcolor{myblue} SpatialSplat & 0.5577 & 0.7920 &  25.47 & 87.7k\\
         \bottomrule
    \end{tabular}
    }
    \caption{\textbf{Ablation study on design choices.} The proposed Adaptive Gaussian mechanism and dual-field architecture jointly enable the construction of high-quality representations with significantly reduced dimensionality.}
\vspace{-1.2em}
\label{table:Ablations}
\end{table}

\vspace{2pt}\noindent\textbf{Effect of selective Gaussian mechanism.}
To answer Question 1, we remove the SGM and retrain our model using the same configuration. As shown in Tab.~\ref{table:Ablations}, the SGM eliminates approximately 35$\%$ of Gaussian primitives while causing only a 0.08 drop in PSNR, indicating that the removed primitives have minimal impact on rendering quality.
The visualization in Fig.~\ref{fig:ablation} (a) and (b) further demonstrates that the SGM accurately localizes overlapping areas in the input images.

\vspace{2pt}\noindent\textbf{Effect of dual-field architecture.}
To address Question 2, we replace our dual-field architecture with single-level per-primitive semantic learning. Following the approach in LSM, we set the semantic feature dimension to 64 and use a convolutional layer to expand it to 512.
As shown in Tab.~\ref{table:Ablations}, this results in a significant drop of 5$\%$ in mIoU for open-vocabulary segmentation performance. 
The primary issue is that per-primitive semantic learning struggles to maintain accurate semantics and fails to preserve clear instance boundaries, as illustrated in Fig.~\ref{fig:ablation} (c) and (d).

\vspace{2pt}\noindent\textbf{Extend to more views.}
Our experiments primarily focus on two-view settings to ensure a fair comparison with our baseline. 
However, our method can be extended to more views for reconstructing larger scenes. 
As shown in Tab.~\ref{table:Ablations}, our approach remains effective even with three input views. 
Moreover, compared to pixel-wise methods, which require approximately 197K primitives to represent the scene, our method achieves a more compact representation by reducing about 50$\%$ primitives.
More qualitative details can be found in the appendix.

%% file: sec/5_conclusion.tex
\section{Conclusion}

\label{sec:Conclusion}
We present an efficient feed-forward network that generates compact yet high-performance representations for 3D reconstruction and scene understanding from sparse, unposed images.
Our proposed dual-field architecture achieves state-of-the-art performance while significantly reducing the number of representation parameters.
Additionally, we introduce a Selective Gaussian Mechanism that directly identifies redundant Gaussians caused by pixel-wise predictions without requiring geometric priors.
SpatialSplat achieves impressive results in 3D reconstruction and scene understanding without relying on any 3D data during training or inference, making semantic 3D reconstruction from sparse, unposed images more practical.
% However, our approach still has some limitations. 
% While effective for static scene reconstruction, the current framework lacks inherent design for long sequential mapping tasks. 
% This presents a promising research direction, as integrating temporal coherence modules could enable applications in SLAM systems and dynamic environment modeling.

%% file: sec/Acknowledgments.tex
\section*{Acknowledgments}

This work was partially supported by Hunan Province Major Scientific and Technological Project No. 2024QK2001, National Key R\&D Program of China No. 2023YFB4704500. \\